\documentclass[10pt]{article} 

\usepackage[accepted]{rlj} 

\usepackage{amssymb}            
\usepackage{mathtools}          
\usepackage{mathrsfs}           
\usepackage{graphicx}           
\usepackage{subcaption}         
\usepackage[space]{grffile}     
\usepackage{url}                
\usepackage{lipsum}             

\usepackage{nicefrac} 
\usepackage{wrapfig} 
\usepackage[capitalize,noabbrev]{cleveref} 
\usepackage{listings} 
\definecolor{listingkeyword}{HTML}{005CC5}
\definecolor{listingcomment}{HTML}{22863A}
\definecolor{listingstring}{HTML}{B31D28}
\definecolor{listingemph}{HTML}{6F42C1}
\crefname{lstlisting}{listing}{listings}
\Crefname{lstlisting}{Listing}{Listings}

\definecolor{aqua}{HTML}{91daf2}
\definecolor{orge}{HTML}{edcd8a}
\definecolor{gren}{HTML}{c6eb8a}
\definecolor{myblue}{HTML}{598BE7}
\definecolor{rliableolive}{HTML}{BBCC33}

\tcbuselibrary{raster,skins,breakable,theorems}
\definecolor{mycitecolor}{HTML}{3498DC}
\definecolor{mylinkcolor}{HTML}{E74D3B}
\definecolor{myurlcolor}{HTML}{980000}
\definecolor{mydarkgreen}{HTML}{6a994e}
\definecolor{myorange}{HTML}{E7730D}
\definecolor{myblue}{HTML}{598BE7}

\newcommand{\contributionbox}[1]{%
\begin{tcolorbox}[
    enhanced,
    colback=myblue!8!white,
    colframe=myblue,
    leftrule=2mm,
    rightrule=0mm,
    toprule=0mm,
    bottomrule=0mm,
    arc=0mm,
    left=5pt,
    right=5pt,
    top=5pt,
    bottom=5pt,
    breakable,
    leftlower=2mm,
    leftupper=2mm
]
\normalsize 
#1
\end{tcolorbox}
}

\crefname{figure}{Figure}{Figures}
\crefname{table}{Table}{Tables}
\crefname{equation}{Equation}{Equations}
\crefname{appendix}{Appendix}{Appendices}
\crefname{section}{Section}{Sections}
\crefname{subsection}{Subsection}{Subsections}

\title{Learning the Supports for Categorical Critic in Reinforcement Learning}

\setrunningtitle{Learning the Supports for Categorical Critic in Reinforcement Learning}

\author{
Jen-Yen Chang \textsuperscript{1,2}, 
Takayuki Osa \textsuperscript{2}, 
Tatsuya Harada \textsuperscript{1,2}
}

\emails{
chou@mi.t.u-tokyo.ac.jp,
takayuki.osa@riken.jp, 
harada@mi.t.u-tokyo.ac.jp
}

\affiliations{
$^{1}$\textbf{Graduate School of Information Science and Technology, The University of Tokyo}\\
$^{2}$\textbf{RIKEN AIP}\\
}

\contribution{We demonstrate that the mean-squared Bellman error, commonly used in RL, is upper-bounded by HL-Gauss, a classification-based value-learning loss.}{The tightness of this upper bound is directly correlated with the absolute values of the support interval. This theoretical connection motivates our approach to actively seek the narrowest feasible support interval to tighten the bound.
}

\contribution{We provide a novel insight that eliminates the reliance on pre-defined support intervals by formulating dynamic support learning as a constrained optimisation problem.}{We discovered that dynamically learning supports requires balancing two opposing forces: minimising the tightness of the upper-bound of the mean squared error to HL-Gauss, whilst simultaneously maximising the truncated Gaussian coverage constraint.
}

\contribution{We formulate the constrained optimisation problem as an adversarial min-max game to derive a tractable learning objective.}{
By introducing a Lagrangian multiplier to enforce a probability mass coverage constraint, we allow the neural network to automatically expand the supports to capture the target distribution's leaked mass, whilst penalising excessive width.
}
 
\contribution{We evaluate the proposed dynamic support learning objective on a range of continuous control tasks.}{Our results show that dynamic support learning is competitive with the fixed-support HL-Gauss baseline on most tasks and improves over it on several tasks.}

\contribution{We provide evidence that a dynamically adapted support can outperform any fixed support.}{We show that no single fixed support is optimal across all policies encountered during training, motivating adaptation over a hand-picked interval.}


\summary{
Value functions are an essential component in actor-critic based deep reinforcement learning (RL). Conventionally, these functions are trained as a regression task by minimising the mean squared error (MSE) relative to bootstrapped target values. Meanwhile, in distributional RL, a distribution of returns is modelled based on the distributional Bellman operator. This work investigates the Gaussian Histogram Loss (HL-Gauss), a recent approach that reframes value estimation as classification by encoding each scalar Bellman target as a Gaussian-smoothed categorical target. Despite its potential, applying histogram-based losses to RL presents inherent challenges, most notably the requirement to pre-define a fixed support interval, which is often complicated by the non-stationary and stochastic nature of target values typically found in RL tasks. In this work, we propose an approach that dynamically learns the lower and upper bounds of the support instead of assigning them beforehand. We derive an objective that jointly learns these bounds whilst learning the categorical representation of the scalar values, and we show that this objective forms an upper bound on the mean-squared Bellman error. Our theoretical analysis further shows that this bound is tighter than that of fixed supports of HL-Gauss. Empirically, the proposed objective enables stable adaptation of the support interval and matches HL-Gauss-based actor-critic algorithms on most continuous-control tasks whilst improving on a subset, without requiring a pre-specified support interval.
}

\begin{document}

\makeCover  
\maketitle  

\begin{abstract}

Value functions are an essential component in actor-critic based deep reinforcement learning (RL). Conventionally, these functions are trained as a regression task by minimising the mean squared error (MSE) relative to bootstrapped target values. Meanwhile, in distributional RL, a distribution of returns is modelled based on the distributional Bellman operator. This work investigates the Gaussian Histogram Loss (HL-Gauss), a recent approach that reframes value estimation as classification by encoding each scalar Bellman target as a Gaussian-smoothed categorical target. Despite its potential, applying histogram-based losses to RL presents inherent challenges, most notably the requirement to pre-define a fixed support interval, which is often complicated by the non-stationary and stochastic nature of target values typically found in RL tasks. In this work, we propose an approach that dynamically learns the lower and upper bounds of the support instead of assigning them beforehand. We derive an objective that jointly learns these bounds whilst learning the categorical representation of the scalar values, and we show that this objective forms an upper bound on the mean-squared Bellman error. Our theoretical analysis further shows that this bound is tighter than that of non-learned supports of HL-Gauss. Empirically, the proposed objective enables stable adaptation of the support interval and matches HL-Gauss-based actor-critic algorithms on most continuous-control tasks whilst improving on a subset, without requiring a pre-specified support interval.

\end{abstract}


\section{Introduction}

Accurate value-function estimation remains a cornerstone of deep reinforcement learning (RL). Traditionally, value functions are approximated through scalar regression using mean-squared error (MSE) with the Bellman update \citep{fujimoto2018td3, haarnoja2018sac}. By minimising the scalar error, the network parameters are updated to drive the point estimate towards the empirical mean of the target. Rather than projecting the Bellman update onto a scalar mean, distributional reinforcement learning (distributional RL) \citep{morimura2010parametric, morimura2010nonparametric, bellemare2017distributional, dabney2018iqn, dabney2018quantile, yang2019fqf, rowland2019statistics,  kuznetsov2020tqc, sun2024sinkhorn} aims to learn the full probability distribution of the random return variable. This paradigm shift fundamentally alters the learning objective from minimising scalar error to minimising the discrepancy between probability distributions, employing metrics such as the Wasserstein distance for quantile-based regression methods or cross-entropy loss for categorical approximations.

Recent literature indicates that framing regression learning as a classification problem, specifically utilising the Gaussian Histogram Loss (HL-Gauss), yields smoother gradients in common regression tasks \citep{imani2018improving, imani2024investigatinghistogramlossregression} and offers superior scalability in RL tasks \citep{stopregressing}. Despite these advantages, histogram-based value-learning methods share a fundamental limitation: the necessity of a pre-defined support interval $[\nu_\text{min}, \nu_\text{max}]$. In practice, however, the true range of expected returns is rarely known \textit{a priori}. This uncertainty is especially pronounced in environments characterised by sparse rewards, unbounded state spaces, or highly non-stationary returns as the policy evolves. Crucially, the return magnitudes induced by early exploratory policies and by the final converged policy can differ substantially, so no single fixed interval is well-suited to every policy encountered during learning.

Consequently, specifying the support interval beforehand introduces an inherent trade-off. Should the pre-defined support interval be excessively narrow, the target distribution will inevitably suffer from boundary truncation. Such truncation discards critical information regarding extreme, yet potentially highly informative, returns, potentially inducing suboptimal policy convergence. Conversely, if the support interval is conservatively set excessively broad to encompass all conceivable returns, the fixed number of bins $k$ must span a vastly broad domain. This dilution reduces the
representational resolution of each bin, weakening the learning signal.

To overcome these limitations, this work introduces a mechanism to dynamically learn the support interval for categorical critics to form the Dynamic Support Endpoint Learning (DySEL) algorithm. The crux of DySEL lies in framing this goal as a constrained optimisation problem. Specifically, the objective is to minimise the width of the learnt support interval in order to maintain high resolution, whilst ensuring it remains sufficiently broad to encapsulate the target distribution, mitigating severe truncation. Empirically, we find that learning the support is competitive with the fixed-support HL-Gauss baseline across most tasks and yields clear gains on a subset of tasks. In summary, the primary contributions of this work are as follows:

\vspace{-1mm}
\contributionbox{
\begin{itemize}
\item We demonstrate that the mean-squared Bellman error, commonly used in RL, is upper-bounded by HL-Gauss, used for classification-based value learning in RL. The tightness of this upper bound is directly correlated with the absolute values of the support interval. This theoretical connection motivates our approach to actively seek the narrowest feasible support interval to tighten the bound.
\item We provide a novel insight that eliminates the reliance on pre-defined support intervals in categorical critics by formulating dynamic support learning as a constrained optimisation problem. We discovered that dynamically learning supports requires balancing two opposing forces: minimising the tightness of the upper-bound of the mean squared error to HL-Gauss, whilst simultaneously maximising the truncated Gaussian distribution coverage.
\item We formulate the constrained optimisation problem as an adversarial min-max game to derive a tractable learning objective. By introducing a Lagrangian multiplier to enforce a probability mass coverage constraint, we allow the neural network to automatically expand the supports to capture the target distribution's leaked mass, whilst penalising excessive width.
\item We evaluate the proposed dynamic support learning objective on a range of continuous control tasks. Our results show that dynamic support learning is competitive with the fixed-support HL-Gauss baseline on most tasks and improves over it on several, most notably the humanoid tasks.
\item We provide evidence that a dynamically adapted support can outperform any fixed support, where no single fixed support is optimal across all policies encountered during training, motivating adaptation over a hand-picked interval.
\end{itemize}
}


\section{Preliminaries}
\label{sec:preliminaries}

A standard RL problem is defined as an infinite-horizon Markov Decision Process (MDP) = ⟨$\mathcal{S}$, $\mathcal{A}$, $P$, $\mathcal{R}$, $\gamma$⟩, where the RL agent at time $t$ observes a state $s_t$ from a set of states $\mathcal{S}$, chooses an action $a$ from a set of actions $\mathcal{A}$, and receives a reward $r$ according to a mapping of the reward function $\mathcal{R}$, $r: \mathcal{S} \times \mathcal{A} \rightarrow \mathbb{R}$. The environment then transitions into a state $s_{t+1}$ with a transition probability function $P(s_{t+1}|s_t, a_t)$ and the interaction continues. We also define the replay buffer $\mathcal{D}$ containing the state, action, reward, and next state at timestep t as $\mathcal{D} = (s_{t}, a_{t}, r_{t}, s_{t+1})$. The objective of an RL agent is to find a policy $\pi$ that maximises the discounted expected return $\mathop{\mathbb{E}}_\pi[\sum_{t=0}^\infty \gamma_t \mathcal{R}(s_{t}, a_{t})]$, which is the expected cumulative sum of rewards when following the policy in the MDP, with the importance of the horizon is determined by a discount factor $\gamma \in [0,1)$. 

\textbf{Distributional Reinforcement Learning.} Whilst standard RL is principally concerned with the expected value of the return $Q^\pi(s, a) = \mathbb{E}[\mathcal{Z}^\pi(s, a)]$, distributional reinforcement learning (distributional RL) \citep{morimura2010parametric, morimura2010nonparametric} seeks to characterise the full probability distribution of the random return variable, denoted as $\mathcal{Z}^\pi(s, a) = \sum_{t=0}^\infty \gamma_t \mathcal{R}(s_t, a_t)$. The distributional RL perspective allows for learning the intrinsic randomness of the environment and the policy, also called aleatoric uncertainty. 
\begin{wrapfigure}{r}{0.5\textwidth}
\includegraphics[width=\linewidth]{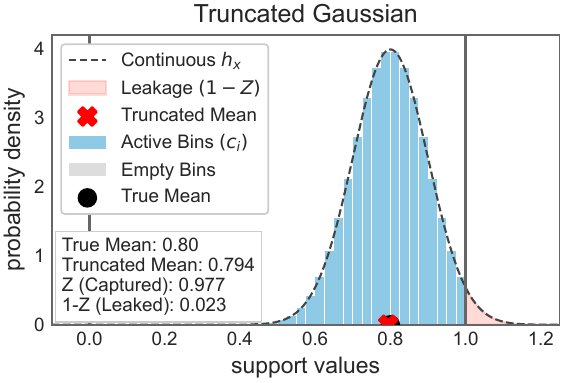}
\vspace{-6mm}
\caption{An example of a truncated Gaussian distribution $q(y)$ and its truncation bias. The X-axis is the support interval, limited in (0,1). The Y-axis is the probability density of the truncated Gaussian. Blue denotes bins with probability assigned, whilst grey bins have zero probability. In this work we aim to utilise $(1-Z)$ as one of the substitutes to learn dynamic support interval.}
\vspace{-7mm}
\label{fig:biases}
\end{wrapfigure}

\textbf{Categorical Value Learning via The Gaussian Histogram Loss (HL-Gauss)}. To represent a categorical distribution within a neural network, it is necessary to discretise the continuous support of the scalar into a fixed set of $k$ atoms or bins. HL-Gauss re-casts the scalar regression target $Q(s,a)$ into a smoothed, continuous probability mass function over the predefined bins, and the neural network is trained to classify the probabilities of each bin by minimising the cross-entropy loss between the output of the neural network $h_i(x)$ and the re-casted target distribution $q_y$. Unlike distributional RL, HL-Gauss does not model the random-return distribution $\mathcal{Z}$. 

Consider a support interval on the real line, denoted by $[\nu_{\text{min}}, \nu_{\text{max}}]$ between the endpoints $\nu_{\text{min}}$ and $\nu_{\text{max}}$, which is uniformly partitioned into $k$ bins, each possessing a width $w_i = (\nu_{\text{max}} - \nu_{\text{min}})/k$. Let $\sigma$ represent the standard deviation, $\mu$ be the actual value of samples, the corresponding truncated Gaussian distribution is defined as:
\begin{equation}
q(y) = \frac{1}{Z\sigma\sqrt{2\pi}} e^{-\frac{(y-\mu)^2}{2\sigma^2}}
\end{equation}
The normalising term $Z$, also referred to in this work as covering mass, is defined as the integral of the probability density function (PDF) of the truncated Gaussian distribution over the support interval, and is unrelated to the random-return variable $\mathcal{Z}^\pi$. Therefore, supposedly a fixed and sufficiently broad support interval is given, $Z \approx 1$. With erf denoting the Gaussian error function \citep{olver2010nist}. $Z$ is calculated as 
\begin{equation}
\label{eq:Z}
Z = \frac{\left( \text{erf}\left(\frac{\nu_{\max} - \mu}{\sqrt{2}\sigma}\right) - \text{erf}\left(\frac{\nu_{\min} - \mu}{\sqrt{2}\sigma}\right) \right)}{2}
\end{equation}
The weighting $c_i$ of i-th bin is defined as 
\begin{align}
c_i &= \frac{1}{2Z}\left(\text{erf}\left(\frac{l_i+w_i-\mu}{\sqrt{2}\sigma}\right) - \text{erf}\left(\frac{l_i-\mu}{\sqrt{2}\sigma}\right)\right)\\
\text{with~} l_i &= \nu_{\min} + (i-1) \cdot w_i ~, \quad i \in \{1, 2, \dots, k\}
\end{align} 
HL-Gauss \citep{imani2018improving, imani2024investigatinghistogramlossregression, stopregressing} is defined as
\begin{equation}
\label{eq:hl-gauss}
\mathcal{L} = -\sum_{i=1}^k c_i \log h_i(x)
\end{equation}
Here $h_i(x)$ is the predicted probability of bin $i$ and $z_i = l_i + w_i/2$ is its centre, so a categorical output decodes to the scalar $\sum_{i=1}^{k} h_i(x)\,z_i$; we write $\mathbb{E}_{q_y}[z]$ for the corresponding decoded mean of the target $q_y$.

\textbf{Biases in HL-Gauss}. When decoding the categorical representation into a scalar value, several biases, namely the truncation bias and the quantisation bias, are inadvertently introduced. In HL-Gauss, these biases are mostly due to the hyperparameter choices. The truncation bias denotes the difference between the mean of a distribution $\mu$ and the mean calculated from the bins as $\mathbb{E}_{q_y}[z]$. That is, $\text{truncation bias} =\mathbb{E}_{q_y}[z] - \mu$. The quantisation bias denotes the difference between the continuous Gaussian $h_x$ and its discretised counterpart onto bins, which is proportional to width $w_i$ of each bin $\text{quantisation bias} \propto w_i = (\nu_\text{max} - \nu_\text{min})/k$ \citep{pmlr-v84-rowland18a, lyle2019comparative}. The quantisation bias also affects the granularity of $c_i$.

Both biases may therefore be viewed as a result introduced by the support interval $[\nu_\text{min}, \nu_\text{max}]$. The truncation bias is prevalent when the support interval is too narrow, resulting in the Gaussian distribution being truncated; whilst the quantisation bias, assuming the number of bins $k$ is fixed, is a direct result of $w_i$, and thus is more prevalent when the support interval is too broad. We show the formulation of both biases as the following and give an intuitive example of a truncated Gaussian distribution and its truncation bias in \cref{fig:biases}, where the sum of the blue part is the covering mass $Z$ calculated from \cref{eq:Z}, from which we may recover the truncated mean by $\mathbb{E}_{q_y}[z]$. The orange part is leaked mass $(1-Z)$, which causes the truncated mean to diverge from the true mean $\mu$. In this work we aim to utilise $(1-Z)$ as one of the substitutes to learn dynamic support interval.


\section{Motivation: Interpretation of the HL-Gauss Loss}
\label{sec:motivation}

Building on the biases of pre-defined support intervals, this section motivates the dynamic adaptation of these intervals for categorical critics in RL. 

We begin by examining the relationship between scalar Q-value regression, typically trained using the mean-squared Bellman error, and classification-based value learning, via HL-Gauss in this instance. Through this analysis, we demonstrate that the mean-squared Bellman error is upper-bounded by the HL-Gauss loss. Crucially, the tightness of this bound correlates directly with the absolute values of the support interval. Thus, enforcing a narrower support yields a theoretically tighter bound on the MSE, motivating our search for the narrowest feasible interval.

\subsection{Connection between HL-Gauss and the upper bound of the mean-squared Bellman error}

Suppose $(\mathcal{T}Q)$ is the Bellman operator, with the neural network outputting samples $(x_j, y_j)$, the target distribution has support interval $[\nu_\text{min}, \nu_\text{max}]$ and target samples are denoted as $(x_j, q_j)$. Extending~\citep{imani2024investigatinghistogramlossregression} to RL, the Bellman operator that minimises with MSE is:
\begin{align}
&\text{MSE}_\text{Bellman} \\
&\coloneq (Q(s,a) - (\mathcal{T}Q)(s,a))^2 \\
&= (Q(s,a) - \mathbb{E}_{q_y}[z] + \mathbb{E}_{q_y}[z] - (\mathcal{T}Q)(s,a))^2 \\
&\leq 2(Q(s,a) - \mathbb{E}_{q_y}[z])^2 + 2(\mathbb{E}_{q_y}[z] - (\mathcal{T}Q)(s,a))^2 \quad \because (M+N)^2 \leq 2M^2 + 2N^2 \\
&= 2(\mathbb{E}_{h_x}[z] - \mathbb{E}_{q_y}[z])^2 + 2(\mathbb{E}_{q_y}[z] - (\mathcal{T}Q)(s,a))^2 \quad \quad \quad \quad \because Q(s,a) := \mathbb{E}_{h_x}[z] \\
&\leq \underbrace{8 \max(|\nu_\text{min}|, |\nu_\text{max}|)^2}_{\text{absolute width}} \cdot \underbrace{\min\left(\frac{1}{2} \mathcal{D}_{KL}(q_y \| h_x), 1 - \exp(-\mathcal{D}_{KL}(q_y \| h_x))\right) \nonumber}_{\text{distributions matching}} \\
& \quad\quad + \underbrace{2(\mathbb{E}_{q_y}[z] - (\mathcal{T}Q)(s,a))^2}_{\text{truncation bias}} \quad \quad \quad\quad \because \text{Equation (7) in \citep{imani2024investigatinghistogramlossregression}} \label{eq:final_bound}
\end{align}
We may observe that the minimisation of the mean-squared Bellman error on the LHS is upper-bounded by the RHS and may be decomposed into three components: the absolute width of the support interval, defined by $[\nu_{\text{min}}, \nu_{\text{max}}]$; the distribution-matching mechanism, facilitated by the Kullback–Leibler (KL) divergence; and the truncation bias, which represents the error between the projected target value and the true target value. 

\subsection{Connection to Previous Works}

In the literature introducing HL-Gauss for both standard regression \citep{imani2018improving, imani2024investigatinghistogramlossregression} and reinforcement learning contexts \citep{stopregressing}, a standard simplifying assumption is that the distributions matching term $\min\left(\frac{1}{2} \mathcal{D}_{KL}(q_y \| h_x), 1 - \exp(-\mathcal{D}_{KL}(q_y \| h_x))\right)$ may be directly substituted with the cross-entropy loss because, during the simultaneous minimisation on both sides (c.f. Section 2.1 in \citep{imani2024investigatinghistogramlossregression}), minimising the cross-entropy effectively minimises the KL divergence $\mathcal{D}_{KL}(q_y \| h_x)$ between the target distribution $q_y$ and the parametrised prediction $h_x$.

Furthermore, existing literature typically assumes a sufficiently broad and fixed support interval $[\nu_\text{min}, \nu_\text{max}]$. Under such conditions, the cross-entropy remains the sole term requiring minimisation. This simplification is theoretically justified; when the support interval is sufficiently wide, the truncation bias, which represents the error incurred by target values falling outside the pre-defined support interval (c.f. \cref{fig:biases}), becomes minimal. That is, the leaked mass $(1-Z)$ of the target probability distribution effectively becomes negligible. Moreover, because the absolute width of the support interval is fixed, it remains constant w.r.t the trainable parameters of the neural network. Consequently, it does not contribute to the gradient computation, allowing prior works to safely discard the absolute width term and focus exclusively on minimising the cross-entropy objective presented in \cref{eq:hl-gauss}.

Nevertheless, since in this work we are interested in dynamic support interval, we can observe in \cref{eq:final_bound} that an excessively broad support interval results in a loose upper bound on the mean-squared Bellman error, resulting in low granularity and potentially leading to suboptimal convergence. Consequently, we contend that it is imperative to dynamically learn an appropriately scaled support interval that balances the resolution of the distribution against the magnitude of the supports. 


\section{Practical Algorithm for Learning Dynamic Supports}
\label{sec:method}

In this work, we present the Dynamic Support Endpoint Learning (DySEL) algorithm derived from the upper bound of the mean-squared Bellman error in \cref{eq:final_bound}. The key to our proposed approach is built on our observation that this upper bound reveals an inherent adversarial tension between the absolute width term $\max(|\nu_\text{min}|, |\nu_\text{max}|)$ and the truncation bias term $\mathbb{E}_{q_y}[z] - (\mathcal{T}Q)(s,a)$ during minimisation. Specifically, the former term, the absolute width, seeks to minimise the support interval directly to have a tighter upper bound on the mean-squared Bellman error, whereas the latter term, the truncation bias, incentivises the expansion of the supports to encapsulate the full probability mass $Z$ in order to eliminate truncation errors. 

To resolve this adversarial tension, we propose to reformulate the objective in \cref{eq:final_bound} as a \textit{constrained optimisation} problem. By making a series of changes, we arrive at our final min-max objective in \cref{eq:minmaxloss} that balances the requirement for a tighter support interval against the necessity of sufficient distributional coverage. 

\textbf{Assumptions}. In the remainder of this work, we operate under the assumption that the number of bins $k$ remains fixed, as it defines the neural network's final output layer, a common assumption in categorical critic literature \citep{bellemare2017distributional,imani2018improving}. Furthermore, in accordance with previous works in HL-Gauss \citep{imani2018improving, imani2024investigatinghistogramlossregression, stopregressing}, the Gaussian standard deviation $\sigma$ is calculated from a pre-determined sigma-to-width ratio, that is, $\sigma \propto w_i$. 
Consequently, the primary remaining degree of freedom for learning is the support interval $[\nu_{\text{min}}, \nu_{\text{max}}]$, which aligns with our goal and the derived upper bound in \cref{eq:final_bound}.

\textbf{Addressing the product.} Firstly, directly minimising the product, in our case $\max(|\nu_\text{min}|, |\nu_\text{max}|) \cdot \text{CE}$ in \cref{eq:final_bound}, is known to cause instability as it constitutes a fundamentally non-convex optimisation problem \citep{boyd2004convex, sun2016nonconvexproblemsscary, zhu2017global, jin2017escape}. Since both the max of absolute values and cross-entropy are always non-negative, we may apply the AM-GM inequality $2\sqrt{X \cdot Y} \le (X + Y)$ and introduce a positive scaling constant $\alpha > 0$ as follows:
\begin{equation}
\label{eq:product}
4 \max(|\nu_\text{min}|, |\nu_\text{max}|)^2 \cdot \text{CE} \le \left( \alpha \max(|\nu_\text{min}|, |\nu_\text{max}|)^2 + \frac{1}{\alpha} \text{CE} \right)^2
\end{equation}
\textbf{Addressing the truncation bias.} Secondly, we argue that minimising the truncation bias may be simplified and transformed into minimising the probability mass residing outside the support interval (leaked mass) $(1-Z)$, which in turn maximises the covering mass $Z$. An intuitive understanding of covering mass $Z$ or leaked mass $(1-Z)$ may be found in \cref{fig:biases}. Let us denote the expected value of this leaked mass as $\mu_{\text{out}}$. By the law of total expectation \citep{weiss2005probability} and after rearranging, 
\begin{align}
(\mathcal{T}Q)(s,a) &= Z \cdot \mathbb{E}_{q_y}[z] + (1 - Z) \cdot \mu_{\text{out}}\\
\mathbb{E}_{q_y}[z] - (\mathcal{T}Q)(s,a) &= (1 - Z)(\mathbb{E}_{q_y}[z] - \mu_{\text{out}})
\end{align}
Since the target is a thin-tailed truncated Gaussian, the displacement $(\mathbb{E}_{q_y}[z] - \mu_{\text{out}})$ is bounded; driving the leaked mass $(1-Z)\to 0$ therefore drives the truncation bias to zero, thus it suffices to control $(1-Z)$ directly.

\textbf{Reformulating as a constrained optimisation problem.} We may now begin to formally present DySEL by observing that the minimisation problem in \cref{eq:final_bound} is at heart a \textit{constrained optimisation} problem. Intuitively, the primary objective is to identify the minimum possible width for the support interval, thereby ensuring a tight upper bound on the error. Simultaneously, this interval must satisfy the constraint of sufficiently encapsulating the probability mass, $Z$, of the truncated Gaussian distribution to prevent detrimental truncation bias. That is, in minimisation terms, we aim to minimise $(1-Z)$. Additionally, one inherent property of $Z$, by definition in \cref{eq:Z}, is that $\text{max}(Z) = 1$, allowing us to utilise $(1-Z)$ as a constraint.

Let $\theta$ denote the parameters of the critic network and $\phi$ represent the parameters of the support interval network. Applying the series of changes presented, we may reformulate the minimisation of the RHS from \cref{eq:final_bound} as the following \textit{constrained optimisation} problem:
\begin{align}
\label{eq:constrainedloss}
\mathcal{J}(\theta, \phi) &= \underbrace{\alpha\max(|\nu_\text{min}|, |\nu_\text{max}|)}_{\text{width penalty}}
+ \underbrace{ \frac{1}{\alpha} (- \sum_{i=1}^{k} c_i \log h_i(x))}_{\text{cross-entropy}} \nonumber \\
\text{subject to~} & \text{the truncation constraint on leaked mass}: (1 - Z) \le \epsilon
\end{align}
\textbf{Our final objective: min-max optimisation.} We are ready to present our final objective. We may optimise the problem in \cref{eq:constrainedloss} as an adversarial min-max game by introducing a Lagrangian multiplier $\lambda$ to enforce the probability mass constraint during the gradient descent process. The parameters $\theta, \phi,$ and $\lambda$ are therefore trained via the following Lagrangian, 
\begin{align}
\label{eq:minmaxloss}
\mathcal{L}(\theta, \phi, \lambda) &= \underbrace{\alpha\max(|\nu_\text{min}|, |\nu_\text{max}|)}_{\text{width penalty}}
+ \underbrace{ \frac{1}{\alpha} (- \sum_{i=1}^{k} c_i \log h_i(x))}_{\text{cross-entropy}}
+ \underbrace{\lambda ((1 - Z) - \epsilon)}_{\text{mass constraint}}
\end{align}
where the optimal parameters $\theta^*$, $\phi^*$ and $\lambda^*$ are defined by:
\begin{align}
\theta^*, \phi^* &= \arg\min_{\theta, \phi} \mathcal{L}(\theta, \phi, \lambda) \quad \text{and} \quad \lambda^* = \arg\max_{\lambda \ge 0} \mathcal{L}(\theta, \phi, \lambda)
\end{align}
Within \cref{eq:minmaxloss}, as $\lambda$ is a trainable parameter, the remaining hyperparameter is $\alpha$, introduced in \cref{eq:product}, which governs the trade-off between the absolute width penalty and the cross-entropy loss when minimising our proposed final objective. Please refer to the \cref{listing:adv:jax} for a reference implementation.


\section{Experiments}
\label{sec:experiments}

In this section, we empirically evaluate the performance of our final proposed objective in \cref{eq:minmaxloss}, comparing it to related HL-Gauss works on a variety of challenging continuous control tasks. We show that our proposed DySEL outperforms or remains competitive with previous baselines. We also provide analyses on some design choices and include ablation studies. Detailed hyperparameters and details on the practical implementation used in our experiments are depicted in \cref{appendix:implement}, with further ablation studies in \cref{appendix:additionalexps}. A reference implementation of HL-Gauss and DySEL is also included in \cref{appendix:implement}.

\subsection{Experimental Setups}

\textbf{Baselines.} For continuous control tasks that we experiment on, we build our algorithm on top of TD3 \citep{fujimoto2018td3}. We also adapt HL-Gauss on top of TD3 to facilitate comparison, noted as TD3+HLG. For all experiments, for TD3+HLG, we use support interval = $[-100, 100]$. We apply DySEL to TD3 and denote the resulting algorithm as TD3+DySEL. Furthermore, for both TD3+HLG and TD3+DySEL, we use $k=128$ bins.

\textbf{Benchmark and Evaluation Method.} We evaluate DySEL on 11 challenging tasks from the commonly used benchmark DeepMind Control Suite \citep{dmcontrol}, where the maximum achievable return for these tasks is 1000. We train 10 seeds, seeds$ = \{0, \ldots 9\}$, for all tasks and train for three million timesteps whilst evaluating every ten thousand timesteps. We run twenty episodes at each evaluation, and calculate inter-quantile mean (IQM) with shaded area as the upper and lower bound of 95\% bootstrapped confidence interval, according to best practices \citep{rliable}. We report the IQM rather than the mean as it is robust to outlier seeds and is the recommended aggregate of \citet{rliable}. The results are depicted in the next subsection. 

\subsection{Results and Q\&As}
Our experiments aim to answer the following questions.

\begin{figure}[t]
\includegraphics[width=\linewidth]{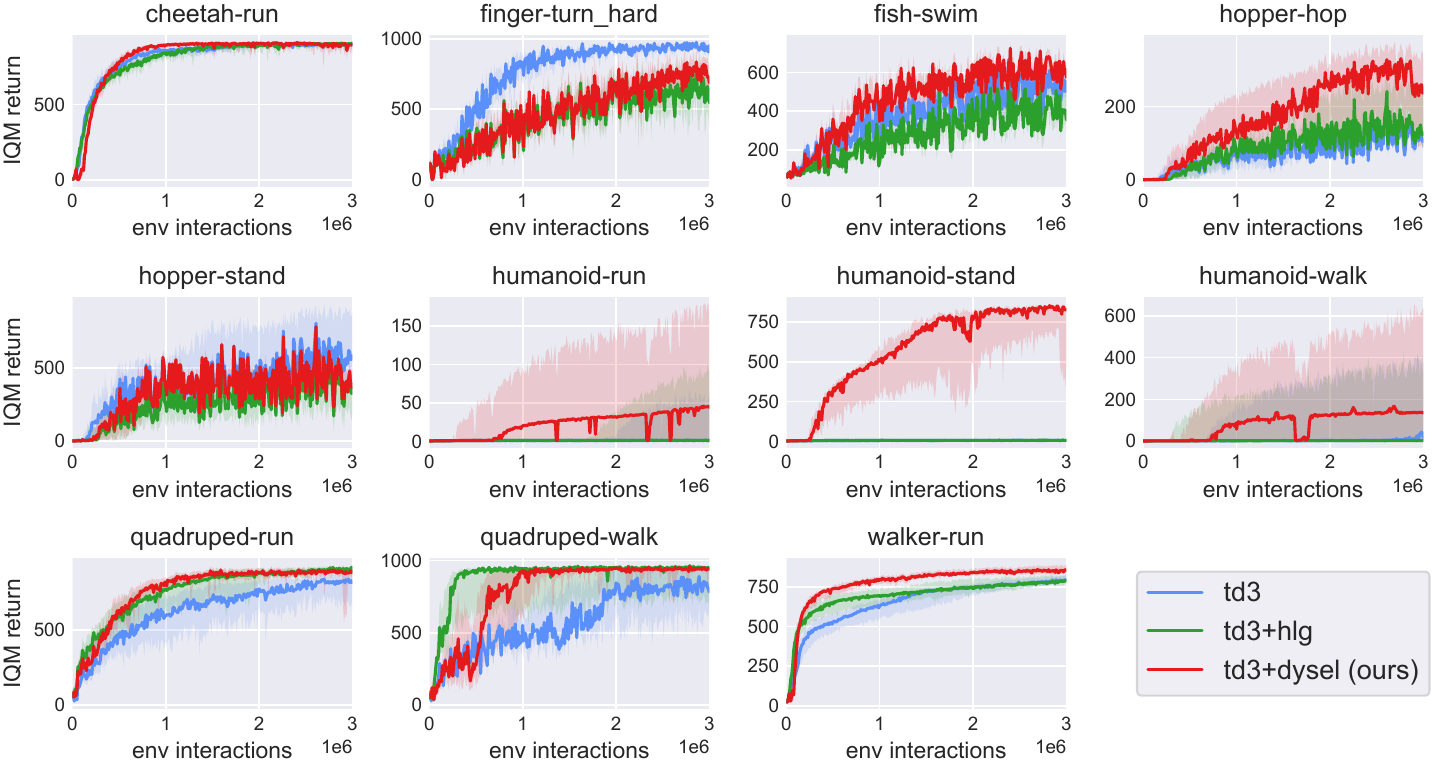}
\vspace{-6mm}
\caption{The IQM returns for each task in DM Control of our proposed DySEL versus baselines. Our proposed method TD3+DySEL remains competitive with TD3+HLG, whilst significantly outperforming in some \textit{humanoid} tasks, showing the empirical benefit of learnt supports.}
\vspace{-6mm}
\label{fig:dmc_results}
\end{figure}

\underline{\textbf{Q1: What is the performance of DySEL for DM-control benchmarking tasks?}} \\
\textbf{A1:} DySEL is generally competitive with vanilla HL-Gauss and provides clear gains on several tasks. 

The IQM returns for each task are depicted in \cref{fig:dmc_results}. Compared with baselines, we find that TD3+DySEL generally achieves better performance or yields results comparable to the best-performing baselines. 

\begin{figure}[t]
\vspace{-0mm}
\includegraphics[width=\linewidth]{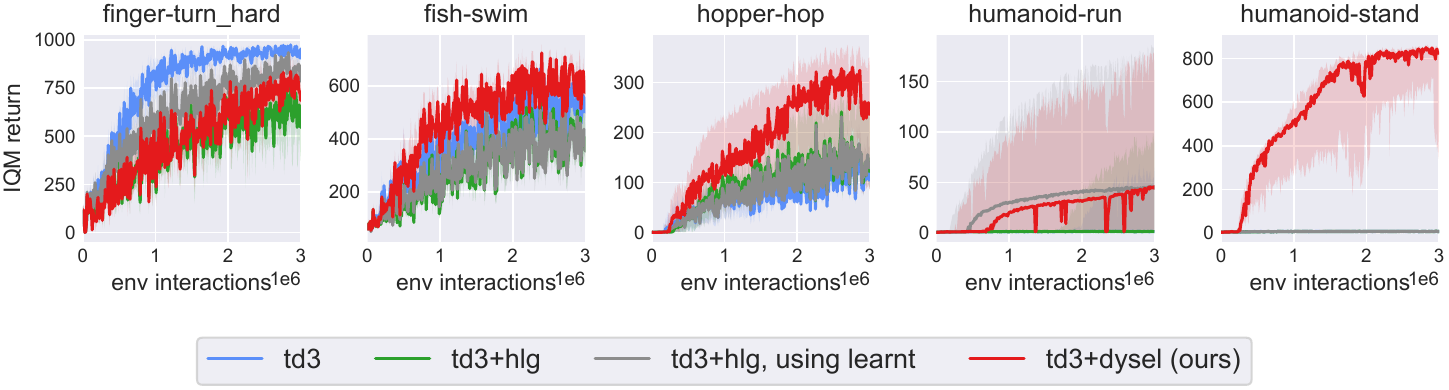}
\vspace{-6mm}
\caption{IQM return comparing DySEL with vanilla HL-Gauss and vanilla HL-Gauss using learnt support interval on selected tasks.}
\vspace{-7mm}
\label{fig:learnt}
\end{figure}

Notably, TD3+DySEL demonstrates substantial performance gains over both TD3 and TD3+HLG for \textit{humanoid} tasks. We hypothesise that this improvement is due to the finer granularity afforded by our dynamically learnt support interval, where the finer resolution enables the critic to discern subtle variations in sample returns, nuances that are typically obscured by pre-defined, fixed supports. Indeed, as depicted in \cref{fig:learnt}, employing these learnt intervals as new reference supports for TD3+HLG yields improved empirical benefits for the \textit{humanoid-run} task. We further discuss using the learnt support interval in Q4.

\begin{figure}[t]
\includegraphics[width=\linewidth]{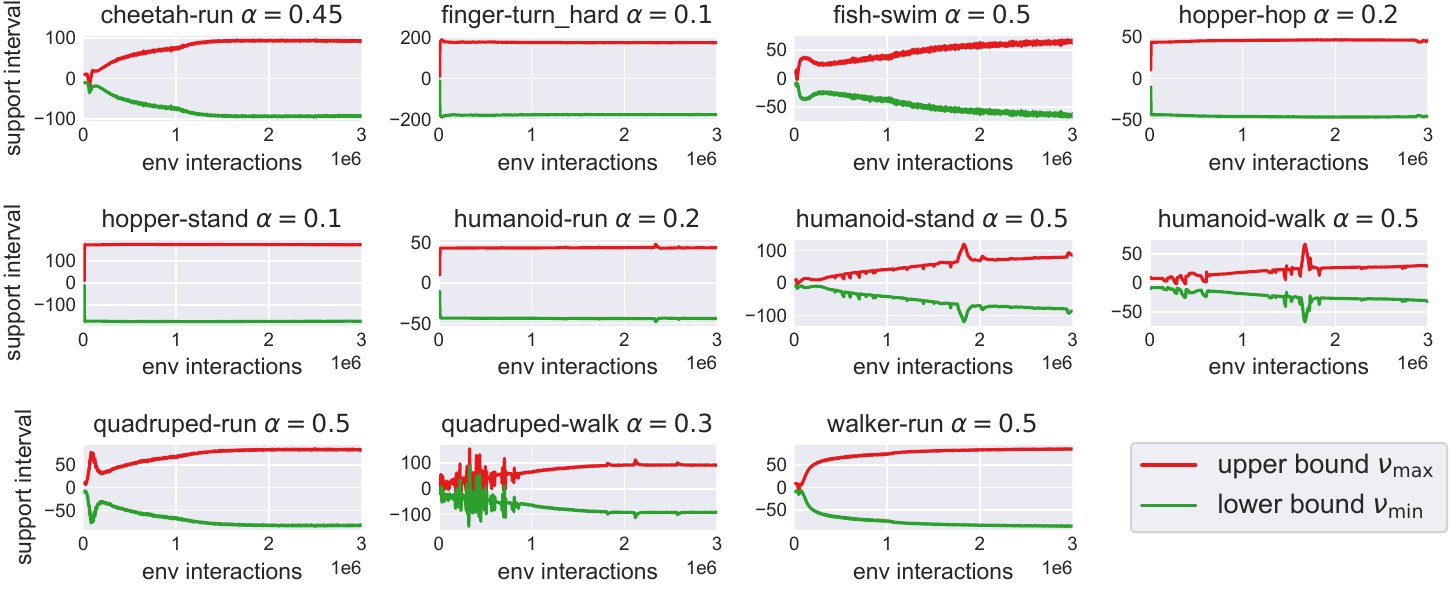}
\vspace{-7mm}
\caption{The evolution of support interval of DySEL during training. We can observe that some support intervals remain stable whilst others gradually increase/decrease as training progresses.}
\vspace{-5mm}
\label{fig:bounds}
\end{figure}

Interestingly, for \textit{finger-turn\_hard} and \textit{hopper-stand} tasks, both HL-Gauss-based approaches either underperform or yield no significant improvement over the baseline TD3 algorithm. Upon examining the evolution of the support intervals for these tasks as depicted in \cref{fig:bounds}, the pattern suggests that the underperformance is tied more to the cross-entropy objective than to dynamic support learning itself, since TD3+HLG also underperforms on these tasks.

\underline{\textbf{Q2: What are the important hyperparameters of DySEL?}} \\
\textbf{A2:} The most important hyperparameter for DySEL is the choice of $\alpha$, which balances between the absolute width penalty and cross-entropy in \cref{eq:minmaxloss}. 

Given that $\alpha$ determines the magnitude of the absolute width penalty, a larger value of $\alpha$ intuitively yields a narrower support interval during the initial stages of training. We observe that empirical experiments corroborate this hypothesis, and we include how these support intervals change during training in \cref{fig:bounds}. Specifically, within the first few thousand steps, $\alpha$ facilitates the rapid learning of a sufficient support interval; subsequently, as training progresses, the learning support interval may gradually become broader, depending on the task and its best $\alpha$. 

Furthermore, we can achieve dynamic $\nu_\text{max}$ and $\nu_\text{min}$ scheduling, such as gradually increasing the support interval, just by tuning the hyperparameter $\alpha$. Instead of manually designing a $\nu_\text{max}$ and $\nu_\text{min}$ schedule, which is often infeasible due to the amount of task-specific knowledge required, by using $\alpha$ we can dynamically learn and expand support interval when the distribution requires more coverage on the covering mass $Z$. 

To be precise, three hyperparameter choices remain. Firstly, there is the important term $\alpha$ between the absolute width penalty and cross-entropy. Secondly, the choice of \textit{initialising} support interval, required to project the first target logits to target scalar values. Thirdly, the choice of number of bins $k$, which is a common hyperparameter for categorical critic methods \citep{bellemare2017distributional, imani2018improving}. Generally speaking, starting from a smaller (e.g. [-10, 10] or [-5, 5]) initialising support interval is a good enough start, but we show different initialising support intervals do not significantly affect performance as detailed in \cref{appendix:additionalexps}. 

\underline{\textbf{Q3: What are the actual values of the supports learnt?}} \\
\textbf{A3:} This is highly task-dependent and closely related to $\alpha$. 

As mentioned previously in Q2, $\alpha$ helps determine the earliest support interval. We include how these support intervals change during training in \cref{fig:bounds}. As we can observe, there are generally speaking two types of behaviours. For some tasks, such as \textit{finger-turn\_hard}, \textit{hopper} tasks and etc., the support interval converges to some stable value almost immediately and remains relatively unchanged, whilst for other tasks, such as \textit{cheetah-run}, \textit{fish-swim}, remaining \textit{humanoid} tasks, \textit{quadruped} tasks and etc., the support interval gradually increases as training progresses. The behaviour is highly dependent on the value of $\alpha$. 

Tasks that require a higher $\alpha$ suggest that finer granularity is important during learning, where the finer resolution enables the critic to discern subtle variations in sample returns. Conversely, tasks that require a more stable support interval would need a lower $\alpha$. Another interesting observation is that our learnt support interval is almost symmetric. 

\underline{\textbf{Q4: How about using the learnt support interval as reference for vanilla HL-Gauss?}} \\
\textbf{A4:} Using the learnt support interval is good enough for vanilla HL-Gauss for some tasks where the support interval remains stable, but not good for tasks where the support interval gradually increases.

Since the support interval changes during training, we use the final support interval that DySEL found (i.e. the support interval at 3M steps), and use it as a referenced support interval for vanilla HL-Gauss for some tasks. We pick these tasks from the two behaviour types we mentioned in Q3. The results are shown in \cref{fig:learnt}.

In tasks such as \textit{finger-turn\_hard} and \textit{humanoid-run}, utilising our learnt support intervals enhances the performance of vanilla HL-Gauss. Whilst this indicates that our learnt intervals provide a robust reference for some environments, this falls short in some tasks. Specifically, for tasks where it is more advantageous to allow the support interval to gradually increase over time, namely in \textit{fish-swim} and \textit{humanoid-stand}, DySEL outperforms or remains competitive vanilla HL-Gauss even when using our learnt support interval. We show that DySEL is able to adapt to both behaviours and to find a balanced support interval for training. This highlights a practical advantage of DySEL: it adapts to both regimes and removes the need to tune the support interval.

\underline{\textbf{Q5: How important are the components in the proposed objective?}} \\
\textbf{A5:} All components of the proposed objective in \cref{eq:minmaxloss} are important.

\begin{figure}[t]
\includegraphics[width=\linewidth]{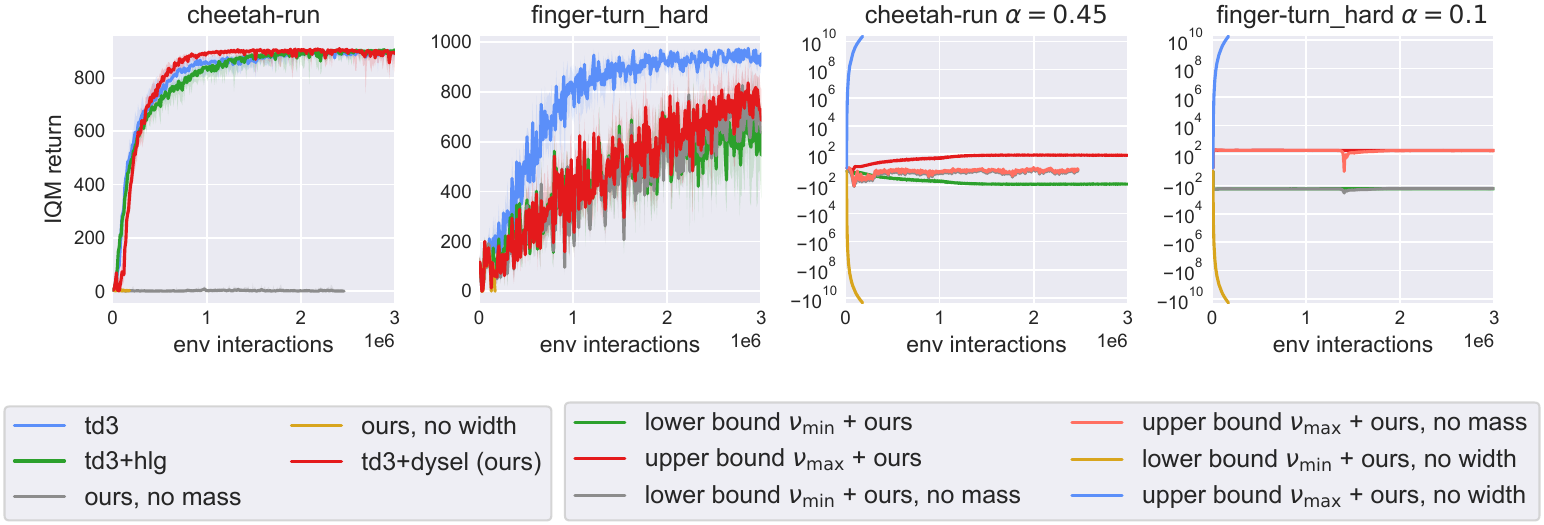}
\vspace{-6mm}
\caption{IQM return and its support interval evolution whilst removing components. No-mass means we remove the mass enforcing adversarial penalty. No-width means we remove the absolute width penalty. In both cases, we can show that both elements are critical for stability. For support interval plots on the right, y-axis is in sym-log scale, because the support interval goes to exceedingly large values and diverges without the width penalty.}
\vspace{-6mm}
\label{fig:fails}
\end{figure}

Whilst \cref{sec:method} outlines the theoretical justification for these components, this section empirically investigates their individual necessity. To this end, we conduct an ablation study examining two specific scenarios: firstly, the removal of the absolute width penalty, thereby allowing the support interval to expand unrestrictedly; and secondly, the omission of the mass constraint, which reduces the constrained optimisation to an unconstrained problem. Empirical results demonstrate that both elements are critical for stability. For both ablation experiments, we maintain the same balancing coefficient, $\alpha$, as utilised in Q1. The corresponding results are illustrated in \cref{fig:fails}.

Observing the results in \cref{fig:fails}, in the absence of the absolute width penalty, the algorithm diverges. We hypothesise that this is because, rather than learning the target distribution, the cross-entropy loss artificially minimises its objective by exploiting the supports (i.e. the tendency to expand $Z$ instead of truly matching the distribution), expanding the support interval towards infinity as a way to minimise its loss. That is, it artificially minimises the $c_i$ parameter in \cref{eq:hl-gauss} by having a larger $Z$ coverage, which in turn affects the learnt support interval by making $|\nu_\text{min}|, |\nu_\text{max}|$ exceedingly large. 

Without the Lagrangian penalty enforcing sufficient mass coverage, the natural tendency of the cross-entropy term to arbitrarily widen the support interval is vastly overpowered by the absolute width penalty. This is acceptable for some tasks such as \textit{finger-turn\_hard}, where the supports remained relatively unchanged (i.e. smaller $\alpha$) in \cref{fig:bounds}. However, for tasks where further expansion of the support interval was necessary, such as \textit{cheetah-run}, the absence of the adversarial penalty's expansion mechanism would result in a support interval too narrow to cover the full target distribution, resulting in insufficient probability mass coverage at some point during training. 

Intuitively, value targets effectively lie in a bounded interval with probability one; increasing the parametrisation support with a fixed bin size eventually places all targets in a single bin, which trivially yields zero cross-entropy. The width penalty is what prevents this degenerate widening.


\section{Related Works}

\textbf{Distributional RL and Classification-Based Value Learning.} The conceptual groundwork for distributional RL was established by \citet{morimura2010parametric,morimura2010nonparametric}, who explored both parametric \citep{morimura2010parametric} and non-parametric \citep{morimura2010nonparametric} approaches. Indeed, distributional RL has been shown to be a critical component in modern state-of-the-art RL algorithms such as BRO \citep{bro} and SimbaV2 \citep{simbav2}. The seminal work on C51 \citep{bellemare2017distributional} introduced the categorical DQN, which discretises the return distribution over a fixed set of atoms or bins. Despite their success, the reliance on a pre-defined support interval introduces a significant dependency that cannot be readily determined \textit{a priori}. On the other hand, HL-Gauss, proposed in \citep{imani2018improving, imani2024investigatinghistogramlossregression}, reframes the regression task as a classification problem by smoothing target values into a Gaussian distribution over a histogram and was adopted to RL target values in \citep{stopregressing}. However, the efficacy of HL-Gauss in RL is still constrained by the requirement of a static support interval. A closely related encoding is the two-hot representation \citep{stopregressing}, which places target mass on the two nearest bins; HL-Gauss can be seen as its Gaussian-smoothed generalisation. In this work, we propose to extend the HL-Gauss framework and learn the support interval dynamically. 

\textbf{Dynamic Supports.} The most related work to DySEL is the work in \citet{chen2025adaptive}, which also has attempted to derive a principled objective based on the upper bound of the mean-squared Bellman error. Whilst they opted for a pure minimisation approach, our approach significantly diverges from theirs where we turn the objective into a constrained optimisation problem, ensuring that the support expands to cover the mass whilst contracting to maintain a tight theoretical bound.

\textbf{Non-Stationarity and Value Scaling.} A primary challenge in RL is the non-stationarity of target values. As the policy improves, the distribution of returns typically shifts and expands, rendering a fixed support interval either too narrow, leading to truncation, or too broad, reducing the resolution per bin. Traditional methods have addressed value scaling through techniques such as PopArt \citep{poparts1, poparts2}, which uses mean-side statistics to normalise targets. Whilst PopArt is effective for scalar regression, it does not naturally extend to the preservation of distributional shape in categorical critics.

\textbf{Quantile Regression.} Implicit Quantile Networks (IQN) \citep{dabney2018iqn} and Fully Parameterized Quantile Function (FQF) \citep{yang2019fqf} avoid using categorical supports completely by learning quantile locations directly, relieving the practitioner from specifying a support range at the price of tail quantiles being hard to estimate. Whilst these types of approaches seem superior as they avoid using fixed pre-defined supports, they suffer from quantile crossing, which may lead to invalid distributions \citep{noncrossing}, and a lack of inductive bias \citep{bellemare2017distributional, imani2018improving, imani2024investigatinghistogramlossregression,sun2025intrinsic}, which may lead to poor performance. DySEL focuses on the categorical-critic branch by making HL-Gauss supports adaptive.

\section{Limitations and Conclusion}

The dynamic adaptation of support interval for categorical critics poses a significant challenge for classification-based categorical value learning~\citep{stopregressing,imani2018improving,imani2024investigatinghistogramlossregression}. Despite the advantages of histogram-based losses, their reliance on a pre-defined and fixed support interval remains problematic when confronted with the non-stationary and stochastic target values inherent to RL tasks, partially due to the amount of hyperparameter tuning required. 

In this work, we propose the Dynamic Support Endpoint Learning (DySEL) algorithm, where we take the first steps towards learning the support interval dynamically. We link the standard mean-squared Bellman error and HL-Gauss to reformulate the learning of support interval from a pure minimisation task into a constrained optimisation problem. By structuring it as a min-max game, the primary objective compresses the support whilst the Lagrangian adversary safeguards sufficient mass to mitigate truncation. Our approach performs comparably to vanilla HL-Gauss on most tasks and improves over it on several, most notably the \textit{humanoid} tasks, whilst eliminating the need to pre-specify the support interval.

Whilst DySEL successfully learns dynamic supports for classification-based categorical value learning, several limitations remain. Notably, the theoretical guarantee of our proposed optimisation relies on a minimax Lagrangian game to enforce the $(1-Z)$ truncation constraint. Although this automates the penalty weight $\lambda$, the stability of the saddle-point optimisation is sensitive to the dual learning rate. Aggressive dual updates can induce oscillatory behaviour in the bounds network during early training phases. Replacing plain dual ascent with a PID-controlled Lagrangian update \citep{stooke2020pid} could damp these oscillations. Another interesting limitation is that our learnt support interval is almost symmetric. Penalising each bound separately could decouple the two and allow an asymmetric shifted support better matched to the true return range, which we leave to future work. Another potential future direction is exploring adaptive Gaussian $\sigma$, allowing finer control over support compression and expansion, and naturally accommodating increasing-$\gamma$ schedules \citep{francoislavet2015discount}. Finally, our evaluation is DM-Control-focused; broader benchmarks are left to future work.

\subsubsection*{Acknowledgments}
\label{sec:ack}
This work was supported by JST Moonshot R\&D Grant Number JPMJPS2011 and JSPS KAKENHI Grant Number JP25K03176.

\bibliography{main}
\bibliographystyle{rlj}

\appendix

\beginSupplementaryMaterials


\section{Experiments Implementation Details \& Hyperparameters}
\label{appendix:implement}

Our implementation is done in JAX \citep{jaxgithub}. Specifically, the versions of important libraries we use in our experiments are: JAX 0.6.2 \citep{jaxgithub}, MuJoCo 3.3.7 \citep{mujoco}, DeepMind Control Suite 1.0.34 \citep{dmcontrol} and gym 0.23.1 \citep{gym}. Nevertheless, we do expect similar empirical performance even if the library versions do not exactly follow ours. We also plan to release our source code upon publication. 

\textbf{Shared across all algorithms}. The replay buffer size is set to $10^6$, and the discount factor $\gamma$ is set to 0.99. The target update rate $\tau$ for target network(s) is 0.005. We have initial random collect steps of 10000. To ensure a fair comparison, all methods employ a batch size of 256, and all neural networks used two hidden layers consisting of 256 units each. All methods use ReLU \citep{relu} as activation function. We use Adam \citep{adam} as optimiser for all neural networks with the learning rate set to $0.0003$. 

\textbf{TD3}. We implemented TD3 \citep{fujimoto2018td3} closely following excellent public repositories such as JAXRL \url{https://github.com/ikostrikov/jaxrl} and high-replay-ratio \citep{highreplayratio} \url{https://github.com/proceduralia/high_replay_ratio_continuous_control}. We use the default hyperparameters provided in these implementations.  

\textbf{HL-Gauss}. For HL-Gauss, we follow the implementation listed in the appendix in \citet{stopregressing}. Whilst the base algorithm was DQN in the original work, we adapted the methodology to TD3 for our experiments involving continuous control tasks. For the critic network, the last layer outputs $k$ values. Furthermore, whereas the baseline employs a fixed support interval of $[\nu_{\text{min}}, \nu_{\text{max}}] = [-10, 10]$ for Atari benchmarks \citep{atari}, we expanded this interval to $[\nu_{\text{min}}, \nu_{\text{max}}] = [-100, 100]$ to accommodate the broader return scales inherent to the DeepMind Control Suite. This range is a principled rather than arbitrary choice: DM Control rewards lie in $[0,1]$ per step, so with a discount factor of $\gamma = 0.99$ the discounted return is bounded by $1/(1-\gamma) = 100$. The interval $[-100, 100]$ therefore covers the full range of attainable values and is essentially free of truncation bias for the baseline; the limitation it does incur is resolution, since realised returns for most tasks and policies remain well below this bound, leaving much of the fixed support---and hence the available bin resolution---unused.

Additionally, the original HL-Gauss work utilises $k=51$ bins, following the convention established by C51 \citep{bellemare2017distributional}. Given that this configuration would yield relatively low granularity across our expanded support (equating to a bin width of nearly 4), we increased the resolution by expanding the number of bins to $k=128$. Finally, consistent with the original authors, we apply a sigma-to-bin-width ratio of 0.75 to determine the standard deviation of the truncated Gaussian. A reference implementation of TD3+HL-Gauss can be found in \cref{listing:losses:jax}. 

\textbf{TD3+DySEL}. We implement our support interval network as a standard three-layer MLP, with hidden dims (256, 256) and the final layer outputting two values. The final layer is initialised according to the initialising support interval, which for all tasks we set to $[-10,10]$. The outputted two values were sorted to ensure that the support interval is valid. The first value is always the lower bound $\nu_\text{min}$ and the second value is always $\nu_\text{max}$. For the specific adversary training, we use Adam \citep{adam} with learning rate of $1e-03$. A reference implementation of our proposed min-max optimisation in \cref{eq:minmaxloss} can be found in \cref{listing:adv:jax}. 

Since our support intervals are dynamically learnt, in \cref{eq:minmaxloss} the remaining relevant hyperparameter is $\alpha$ that balances the minimisation of absolute width versus the minimisation of cross-entropy. We use $k=128$ bins and sigma-to-width ratio = 0.75, same as in our HL-Gauss implementation. Furthermore, we use $\epsilon=0.005$ for all tasks for our proposed min-max optimisation \cref{eq:minmaxloss} and did not tune any further.

\textbf{Selection of $\alpha$.} For each task, we selected $\alpha$ via a grid search over the range $[0.1, 0.2, 0.3, 0.4, 0.45, 0.5]$; the chosen per-task values are listed in \cref{table:alpha_hyperparams}. We found this range sufficient across all tasks, with smaller $\alpha$ favouring a stable, narrow support and larger $\alpha$ permitting gradual expansion, as characterised by the sensitivity analysis in \cref{fig:alpha}. All other hyperparameters are shared with the HL-Gauss baseline, so $\alpha$ is the only method-specific quantity tuned; in effect, DySEL replaces the task-specific tuning of the support interval $[\nu_\text{min}, \nu_\text{max}]$ required by HL-Gauss with the tuning of a single scalar $\alpha$. We note that $\alpha$ was selected on the same seeds subsequently used to report results in \cref{fig:dmc_results}; the per-task performance of DySEL should therefore be read as an optimistic (upper-bound) estimate rather than an unbiased one. We report the full sensitivity of performance to $\alpha$ in \cref{fig:alpha} so that the variation across the searched range can be assessed directly.

\begin{table}[!h]
\centering
\caption{Hyperparameter $\alpha$ that balances the weights of absolute width penalty versus the weights of cross-entropy when minimising.}
\begin{tabular}{c||c|c|c|c}
Task             & cheetah-run & finger-turn\_hard & fish-swim & hopper-hop \\ 
$\alpha$         &  0.45 & 0.1 & 0.5 & 0.2   \\ \hline
Task             & hopper-stand & humanoid-run & humanoid-stand & humanoid-walk \\ 
$\alpha$         &  0.1 & 0.2 & 0.5 & 0.5   \\ \hline
Task             &  quadruped-run & quadruped-walk & walker-run & \\
$\alpha$         &  0.5 & 0.3 & 0.5 &        \\

\end{tabular}
\label{table:alpha_hyperparams}
\end{table}

One important implementation detail is that we clip the minimum value of $\sigma$. If we do not perform this clipping, when minimising $(1-Z)$, the network may cheat by making $\sigma$ smaller in the $Z$ formula (restated from \cref{eq:Z} below for convenience) especially at the beginning steps of training. Specifically, we clip the minimum to $\sigma=0.3$.

\begin{equation}
\text{(\text{\cref{eq:Z}} restated)~}Z = \frac{1}{2} \left( \text{erf}\left(\frac{\nu_{\max} - \mu}{\sqrt{2}\sigma}\right) - \text{erf}\left(\frac{\nu_{\min} - \mu}{\sqrt{2}\sigma}\right) \right) \nonumber
\end{equation}

\begin{lstlisting}[
float=!h,
caption={A reference implementation of HL-Gauss \citep{imani2018improving} in JAX.},
label={listing:losses:jax}
]

import jax.numpy as jnp

...
# next q
next_q1_logits, next_q2_logits = target_critic.apply_fn(
    target_critic.params,
    batch.next_observations,
    next_actions,
)
next_q1 = hl_gauss.transform_from_logits_to_value(next_q1_logits)
next_q2 = hl_gauss.transform_from_logits_to_value(next_q2_logits)
next_q = jnp.minimum(next_q1, next_q2)

# target q
target_q = batch.rewards + discount * batch.masks * next_q
target_probs = hl_gauss.transform_from_value_to_probs(target_q)

def critic_loss_fn(trainable_params):
    q1_logits, q2_logits = critic.apply_fn(
        trainable_params,
        batch.observations,
        batch.actions,
    )
    
    # cross entropy
    ce_loss1 = optax.softmax_cross_entropy(q1_logits, target_probs)
    ce_loss2 = optax.softmax_cross_entropy(q2_logits, target_probs)
    critic_loss = (ce_loss1 + ce_loss2).mean()
...

\end{lstlisting}

\begin{lstlisting}[
float=!h,
caption={A reference implementation of our proposed objective in \cref{eq:minmaxloss} in JAX.},
label={listing:adv:jax}
]


import jax.numpy as jnp

...

def critic_loss_fn(trainable_params):
    ...
    
    # widths
    lower, upper = q_bounds[:, 0], q_bounds[:, 1]
    M = jnp.maximum(jnp.abs(lower), jnp.abs(upper))
    first_term = M * bounds_weight + ce_loss / bounds_weight

    # bias penalty
    projection_bias = 1.0 - target_z
    second_term = adv_lambda * jax.nn.relu(projection_bias - epsilon)

    # total
    critic_loss = first_term + second_term
    critic_loss = critic_loss.mean()
    ...

# we only need the (1-Z-epsilon) for training adversarial loss
projection_bias = 1.0 - target_z
violation = projection_bias - epsilon

def adv_loss_fn(trainable_params):
    adv_lambda = adversary.apply_fn(trainable_params)
    loss = -(adv_lambda * violation).mean()

...

\end{lstlisting}


\section{Additional Experiments}
\label{appendix:additionalexps} 

As mentioned in \cref{sec:experiments}, three choices of hyperparameters, the initialising support interval, $\alpha$, and number of bins $k$ remain. We show that $\alpha$ has a non-trivial effect on the training performance for some tasks. Furthermore, we also tried different initialising support intervals for the task, which generally has not much effect apart from some harder tasks.

\subsection{Ablation on the initialising support interval}

For all tasks, we previously used an initialising support interval of $[-10,10]$. Empirically, we found that starting from an excessively large initialising support interval can lead to unstable training. We hypothesise that this is because of absolute width penalty of the proposed objective \cref{eq:minmaxloss}, which would be too large to appropriately learn the cross-entropy loss. Nevertheless, we find the ability to be able to start with arbitrary narrower initialising support interval to be more important in practice. To this end, we compared using three different sets of initialising support interval: $[-1,1]$, $[-5,5]$ and $[-10,10]$ and train on some tasks. The results are shown below. 

We can observe that the support interval evolution is generally the same regardless of the initialising support interval. Only for \textit{humanoid} tasks where initially there is some instability. This instability may encourage the policy to explore more. We leave this for future work.

\begin{figure}[h]
\includegraphics[width=\linewidth]{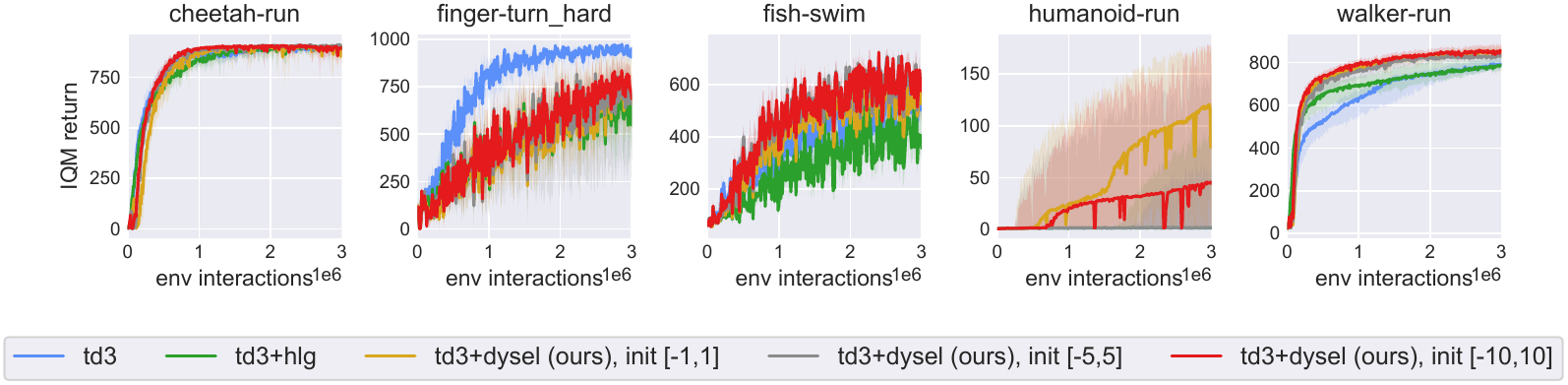}
\includegraphics[width=\linewidth]{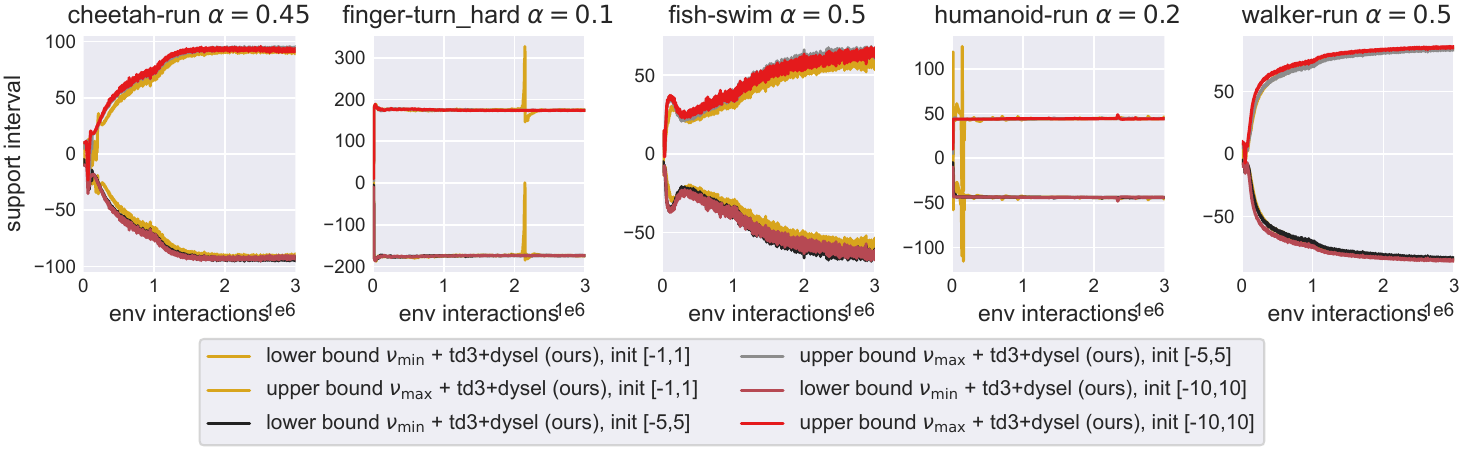}
\caption{IQM and bounds evolution of different initialisation. For all tasks, the supports evolution are similar, but those initialised with smaller initialising supports have some instability.}
\label{fig:init}
\end{figure}

\subsection{Ablation on the sensitivity to $\alpha$}

\begin{figure}[h]
\includegraphics[width=\linewidth]{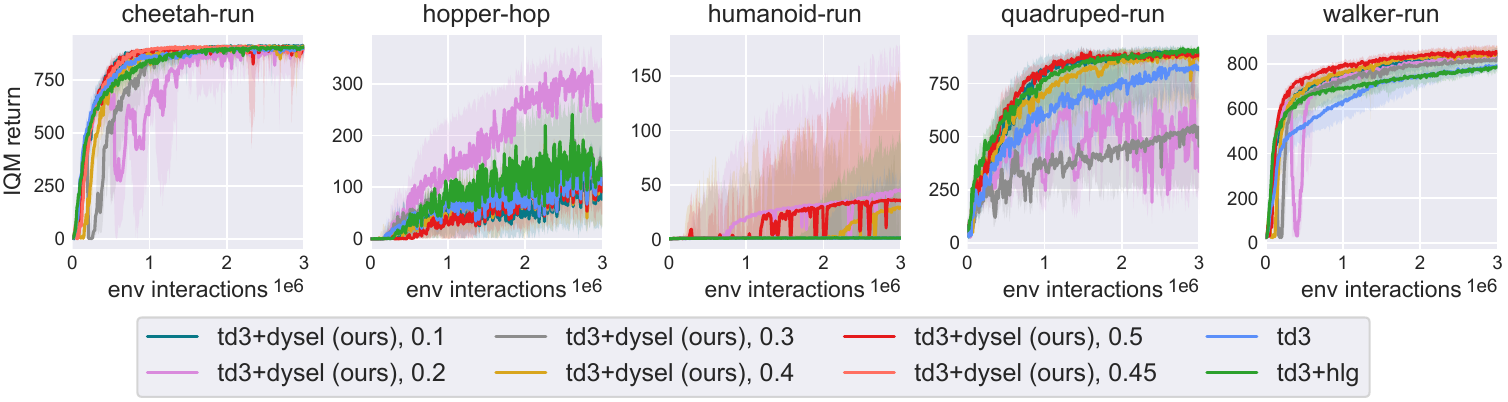}
\caption{IQM and bounds evolution of different $\alpha$.}
\label{fig:alpha}
\end{figure}

For all tasks, we searched for different $\alpha$, which balances between the absolute width penalty and cross-entropy in \cref{eq:minmaxloss}. For some tasks, smaller $\alpha$ is required to maintain a stable support interval; whilst for some tasks where lower granularity is required, larger $\alpha$ is preferred. We include sensitivity to $\alpha$ on some tasks in \cref{fig:alpha}.

\end{document}